\documentclass[11pt,a4paper]{article}
\usepackage[hyperref]{acl2021}
\usepackage{times}
\usepackage{latexsym}

\usepackage{microtype}
\usepackage{graphicx}

\def\argmax{\mathop{\rm argmax}}%
\def\argmin{\mathop{\rm argmin}}%

\usepackage{bm}
\usepackage{algorithm}
\usepackage{algpseudocode}

\usepackage{amsthm,amsmath,amssymb}

\usepackage{mathrsfs}

\aclfinalcopy 

\usepackage{lipsum}

\title{Alternated Training with Synthetic and Authentic Data \\ for Neural Machine Translation}

\author{Rui Jiao$^{1,3,4}$, Zonghan Yang$^{1,3,4}$, Maosong Sun$^{1,3,4,5}$, and Yang Liu\thanks{\quad Corresponding author: Yang Liu}$^{\ \ 1,2,3,4,5}$ \\
  $^1$Department of Computer Science and Technology, Tsinghua University, Beijing, China \\
  $^2$Institute for AI Industry Research, Tsinghua University, Beijing, China \\
  $^3$Institute for Artificial Intelligence, Tsinghua University, Beijing, China \\
  $^4$Beijing National Research Center for Information Science and Technology \\ 
  $^5$Beijing Academy of Artificial Intelligence\\}

\date{}

\begin{document}
\maketitle
\begin{abstract}

While synthetic bilingual corpora have demonstrated their effectiveness in low-resource neural machine translation (NMT), adding more synthetic data often deteriorates translation performance. In this work, we propose alternated training with synthetic and authentic data for NMT. The basic idea is to alternate synthetic and authentic corpora iteratively during training. Compared with previous work, we introduce authentic data as guidance to prevent the training of NMT models from being disturbed by noisy synthetic data.
Experiments on Chinese-English and German-English translation tasks show that our approach improves the performance over several strong baselines.
We visualize the BLEU landscape to further investigate the role of authentic and synthetic data during alternated training. From the visualization, we find that authentic data helps to direct the NMT model parameters towards points with higher BLEU scores and leads to consistent translation performance improvement.

\end{abstract}

\section{Introduction}

While recent years have witnessed the rapid development of Neural Machine Translation (NMT) \citep{NIPS2014_a14ac55a,bahdanau2015neural,gehring2017convolutional,vaswani2017attention}, it heavily relies on large-scale, high-quality bilingual corpora. Due to the expense and scarcity of authentic corpora, synthetic data has played a significant role in boosting translation quality \citep{DBLP:conf/nips/HeXQWYLM16,DBLP:conf/acl/SennrichHB16,DBLP:conf/emnlp/ZhangZ16,DBLP:conf/acl/ChengXHHWSL16,fadaee-etal-2017-data}. 

Existing approaches to synthesizing data in NMT focus on leveraging monolingual data in the training process. Among them, back-translation (BT) \citep{DBLP:conf/acl/SennrichHB16} has been widely used to generate synthetic bilingual corpora by using a trained target-to-source NMT model to translate large-scale target-side monolingual corpora. Such synthetic data can be used to improve source-to-target NMT models. Despite the effectiveness of back-translation, the synthetic data inevitably contains noise and erroneous translations. As a matter of fact, it has been widely observed that while BT is capable of benefiting NMT models by using relatively small-scale synthetic data, further increasing the quantity often deteriorates translation performance \citep{edunov2018understanding,wu2019exploiting,DBLP:conf/wmt/CaswellCG19}. 

This problem has attracted increasing attention in the NMT community \citep{edunov2018understanding,DBLP:conf/emnlp/WangLWLS19}. One direction to alleviate the problem is to add noise or a special tag on the source side of synthetic data, which enables NMT models to distinguish between authentic and synthetic data \citep{edunov2018understanding,DBLP:conf/wmt/CaswellCG19}. Another direction is to filter or evaluate the synthetic data by calculating confidence over corpora, making NMT models better exploit synthetic data \citep{DBLP:conf/aclnmt/ImamuraFS18,DBLP:conf/emnlp/WangLWLS19}. 
While these methods have outperformed the conventional BT approach, NMT models still suffer from a performance degradation as the size of synthetic data keeps increasing. 
Hence, how to better take advantage of limited authentic data and abundant synthetic data still remains a grand challenge.

In this work, we propose alternated training with synthetic and authentic data for neural machine translation. The basic idea is to alternate synthetic and authentic corpora iteratively during training. 
Compared with previous work, we introduce authentic data as guidance to prevent the training of NMT models from being disturbed by noisy synthetic data. Our approach is inspired by the characterization of synthetic and authentic corpora as two types of different approximations for the distribution of infinite authentic data.
We visualize the BLEU landscape to further investigate the role of authentic and synthetic data during alternated training. We find that the authentic data helps to direct NMT model parameters towards the points with higher BLEU scores.
Experiments on Chinese-English translation tasks show that our approach improves the performance over strong baselines.

\section{Alternated Training}



Let $\mathbf{x}$ be a source sentence and $\mathbf{y}$ be a target sentence. We use $P(\mathbf{y}|\mathbf{x}; \bm{\theta})$ to denote an NMT model parameterized by $\bm{\theta}$. Let $D_{a} = \{ \langle \mathbf{x}_{n}, \mathbf{y}_{n} \rangle \}_{n=1}^{N}$ be an authentic parallel corpus containing $N$ sentence pairs. Traditional NMT aims to obtain $\hat{\bm{\theta}}_{a}$ that maximizes the log-likelihood on $D_{a}$:
\begin{equation}
    \hat{\bm{\theta}}_{a} = \argmax_{\bm{\theta}} \Bigg\{ \frac{1}{N}\sum_{n=1}^{N} \log P(\mathbf{y}_n|\mathbf{x}_n; \bm{\theta}) \Bigg\}.
\label{au_eq}
\end{equation}

Back-translation generates additional synthetic parallel data from the monolingual corpus. Let $D_m = \{ \mathbf{y}_{m} \}_{m=1}^{M}$ be a monolingual corpus containing $M$ target-side sentences. Back-translation first trains a target-to-source model  $\hat{\bm{\theta}}_{\mathrm{BT}}$ on $D_{a}$:
\begin{equation}
    \hat{\bm{\theta}}_{\mathrm{BT}} = \argmax_{\bm{\theta}} \Bigg\{ \frac{1}{N}\sum_{n=1}^{N} \log P(\mathbf{x}_n|\mathbf{y}_n; \bm{\theta}) \Bigg\},
\end{equation}
which is then used to translate each sentence in the target-side monolingual corpus $D_m$:
\begin{equation}
\hat{\mathbf{x}}_m = \argmax_{\mathbf{x}} \Big\{ P(\mathbf{x}|\mathbf{y}_{m};\hat{\bm{\theta}}_{\mathrm{BT}}) \Big\},
\end{equation}
where $m=1,\dots,M$. The synthetic corpus $D_{s}$ is generated by pairing the translations $\{ \hat{\mathbf{x}}_{m} \}_{m=1}^{M}$ with $D_m$, i.e. $D_{s} = \{ \langle \hat{\mathbf{x}}_{m}, \mathbf{y}_{m} \rangle \}_{m=1}^{M}$. The required source-to-target model is finally trained on the combination of authentic and synthetic data:
\begin{equation}
    \begin{aligned}
        \hat{\bm{\theta}}_{s}\!=\!\argmax_{\bm{\theta}}\!\Bigg\{\!\frac{1}{N\!+\!M} \Big(& \sum_{n=1}^{N} \log P(\mathbf{y}_n|\mathbf{x}_n; \bm{\theta}) \, + \\
        &\!\sum_{m=1}^{M} \log P(\mathbf{y}_m|\hat{\mathbf{x}}_m; \bm{\theta}) \Big)\! \Bigg\}.
    \end{aligned}
    \label{BT_eq}
\end{equation}


\begin{algorithm}[t] 
\caption{Alternated Training for NMT} 
\label{alg:AlgCL} 
\begin{algorithmic}[1]
\Require Synthetic data $D_{s} = \{ \langle \hat{\mathbf{x}}_{m}, \mathbf{y}_{m} \rangle \}_{m=1}^{M}$, Authentic data $D_{a} = \{ \langle \mathbf{x}_{n}, \mathbf{y}_{n} \rangle \}_{n=1}^{N}$
\Ensure $\hat{\bm{\theta}}_{\mathrm{alter}}$
\State Set $\hat{\bm{\theta}}_{a}^{(0)}$ as random initialization; 
\State $t \leftarrow 0$; 
\State \textbf{while} Not Converged \textbf{do}
\State \indent Obtain $\hat{\bm{\theta}}_{s}^{(t+1)}$ on $D_{s}\cup D_{a}$ with $\hat{\bm{\theta}}_{a}^{(t)}$ as the starting point using Eq. (\ref{BT_eq}); \qquad\,\,\,\, $\vartriangleright$ \textbf{S-Step} 
\State \indent Obtain $\hat{\bm{\theta}}_{a}^{(t+1)}$ on $D_{a}$ with $\hat{\bm{\theta}}_{s}^{(t+1)}$ as the starting point using Eq. (\ref{au_eq}); \qquad\,\,\, $\vartriangleright$ \textbf{A-Step} 
\State \indent $t \leftarrow t+1$; 
\State \textbf{end while}
\State \textbf{return} $\hat{\bm{\theta}}_{\mathrm{alter}} = \hat{\bm{\theta}}_{a}^{(t)}$.
\end{algorithmic}
\end{algorithm}




Suppose that there exists infinite authentic parallel data, which can be characterized as distribution $p(\mathbf{x},\mathbf{y})$. Synthesizing the large-scale corpus $D_s$ is to better approach the authentic parallel data distribution. 
Furthermore, the finite corpora $D_{a}$ and $D_{s}\cup D_{a}$ can be viewed as different empirical approximations of $p(\mathbf{x},\mathbf{y})$:
\vspace{0pt}
\begin{align}
    p_{a}(\mathbf{x},\mathbf{y}) & = \frac{1}{N}\sum_{n=1}^{N} \delta_{\langle\mathbf{x}_n, \mathbf{y}_n\rangle \in D_{a}}(\mathbf{x},\mathbf{y}), \\
    p_{s}(\mathbf{x},\mathbf{y}) & = \frac{1}{N\!+\!M}\Bigg(\, \sum_{n=1}^{N} \delta_{\langle \mathbf{x}_n, \mathbf{y}_n\rangle  \in D_{a}}(\mathbf{x},\mathbf{y}) \, + \nonumber \\
    & \quad\quad\quad\quad\quad\,\, \sum_{m=1}^{M} \delta_{\langle\hat{\mathbf{x}}_m, \mathbf{y}_m\rangle \in D_{s}}(\mathbf{x},\mathbf{y}) \Bigg),
\end{align}
where $\delta$ represents the Dirac distribution. 
On the one hand, $D_a$ is considered to be of higher quality as $\lim_{N \to \infty} p_a(\mathbf{x},\mathbf{y}) = p(\mathbf{x},\mathbf{y})$ exactly recovers the authentic data distribution.
On the other hand, although $D_{s}$ contains certain noise (as $\lim_{M \to \infty} p_s(\mathbf{x},\mathbf{y}) \ne p(\mathbf{x},\mathbf{y})$), it provides more diversified data samples that enable the NMT model to reconstruct the global distribution. As the two corpora are complementary to each other, we introduce authentic data periodically during the training process with synthetic data. Intuitively, alternated training using authentic corpora helps to rectify the deviation of training direction affected by the noisy synthetic data and enhances model performance.

Our proposed alternated training approach is shown in Algorithm~\ref{alg:AlgCL}. 
Starting with random initialization, each alternation cycle during training consists of two steps.
For the $t$-th cycle, the first step is to finetune the model $\hat{\bm{\theta}}_{a}^{(t)}$ with Eq. (\ref{BT_eq}) on $D_{s} \cup D_{a}$ until convergence\footnote{We also attempted to train S/A-steps for certain iterations. Empirically, the proposed convergence-based method performed better.} to obtain $\hat{\bm{\theta}}_{s}^{(t+1)}$, which is referred as \textbf{S-Step} (line 4).
The second step is to alter the training data back to $D_{a}$ and finetune $\hat{\bm{\theta}}_{s}^{(t+1)}$ with Eq. (\ref{au_eq}) until convergence to obtain $\hat{\bm{\theta}}_{a}^{(t+1)}$, which is referred as \textbf{A-Step} (line 5). 
We alternate the training process until convergence. 
It is noted that back-translation is equivalent to a single S-Step performed in our approach.




\begin{table*}[htbp]
\centering
\begin{tabular}{l|l|lllll|l}
\hline
Data & NIST06 & NIST02 & NIST03 & NIST04 & NIST05 & NIST08 & All\\
\hline
\hline
Base & 45.94 & 45.82 & 45.35 & 46.88 & 45.43 & 36.98 & 44.40 \\
BT & 43.89 & 44.79 & 44.40 & 46.24 & 45.45 & 36.45 & 43.57 \\
BT-tagged & 46.79 & 47.11 & 46.49 & 47.73 & 47.17 & 38.41 & 45.47 \\
\hline
AlterBT & 49.07$^{+\dag}$ & 48.77$^{+\dag}$ & 48.36$^{+\dag}$ & \textbf{49.51}$^{+\dag}$ & \textbf{49.94}$^{+\dag}$ & \textbf{40.95}$^{+\dag}$ & \textbf{47.68}$^{+\dag}$ \\
AlterBT-tagged & \textbf{49.40}$^{+\dag}$ & \textbf{49.04}$^{+\dag}$ & \textbf{48.37}$^{+\dag}$ & 49.10$^{+\dag}$ & 49.64$^{+\dag}$ & 40.56$^{+\dag}$ & 47.49$^{+\dag}$ \\
\hline
\end{tabular}
\caption{\label{zh-en-nist}
BLEU scores on the NIST Chinese-English task with 10M additional synthetic corpus. ``Base'' means only authentic data is used. ``BT'' corresponds to the back-translation method \citep{DBLP:conf/acl/SennrichHB16}. ``BT-tagged'' corresponds to the tagged BT technique proposed by \citet{DBLP:conf/wmt/CaswellCG19}. ``AlterBT'' means alternated training on authentic data and synthetic data using ``BT'' in each alternation. ``AlterBT-tagged'' means alternated training on authentic data and synthetic data using ``BT-tagged'' in each alternation. ``$+$" means significantly better than BT (p $<$ 0.01).``\dag" means significantly better than BT-tagged (p $<$ 0.01).}
\end{table*}

\section{Experiments}

\subsection{Setup}


We evaluated our training strategy on Chinese-English and German-English translation tasks. We reported the tokenized BLEU score as calculated by multi-bleu.perl. 

For the Chinese-English task, we extracted 1.25M parallel sentence pairs from LDC as our authentic bilingual corpus and 10M English-side sentences from WMT17 Chinese-English training set as our monolingual corpus for back-translation. NIST06 was used as the validation set. We use NIST02, 03, 04, 05 and 08 datasets as test sets. For the German-English task, we selected the dataset of IWSLT14 German-English task, which contains 16k parallel sentence pairs for training. We further extracted 4.5M English-side sentences from WMT14 German-English training set as monolingual dataset. We segmented Chinese sentences by THULAC \citep{sun2016thulac} and tokenized English and German sentences by Moses \citep{DBLP:conf/acl/KoehnHBCFBCSMZDBCH07}. The vocabulary was built by Byte Pair Encoding (BPE) \citep{DBLP:conf/acl/SennrichHB16a} with 32k merge operations.
We used Transformer \citep{vaswani2017attention} implemented in THUMT \citep{tan-etal-2020-thumt} with standard hyperparameters as a base model. We used Adam optimizer \citep{Adam} with $\beta_1=0.9$, $\beta_2=0.98$ and $\epsilon=10^{-9}$ with 
the maximum learning rate $= 7\times10^{-4}$. 

We applied early-stopping to verify convergence of each single S/A-step.  If the validation BLEU failed ti exceed the highest score during the certain S/A-step after 10K training iterations, we consider the model converged and alternated the training set. For the whole training process, we set the maximum training iterations as 250k for Chinese-English task and 150k for German-English task.

\subsection{Results}

\begin{figure}[t]
\centering
\begin{minipage}{1\columnwidth}
 \includegraphics[width=0.95\columnwidth]{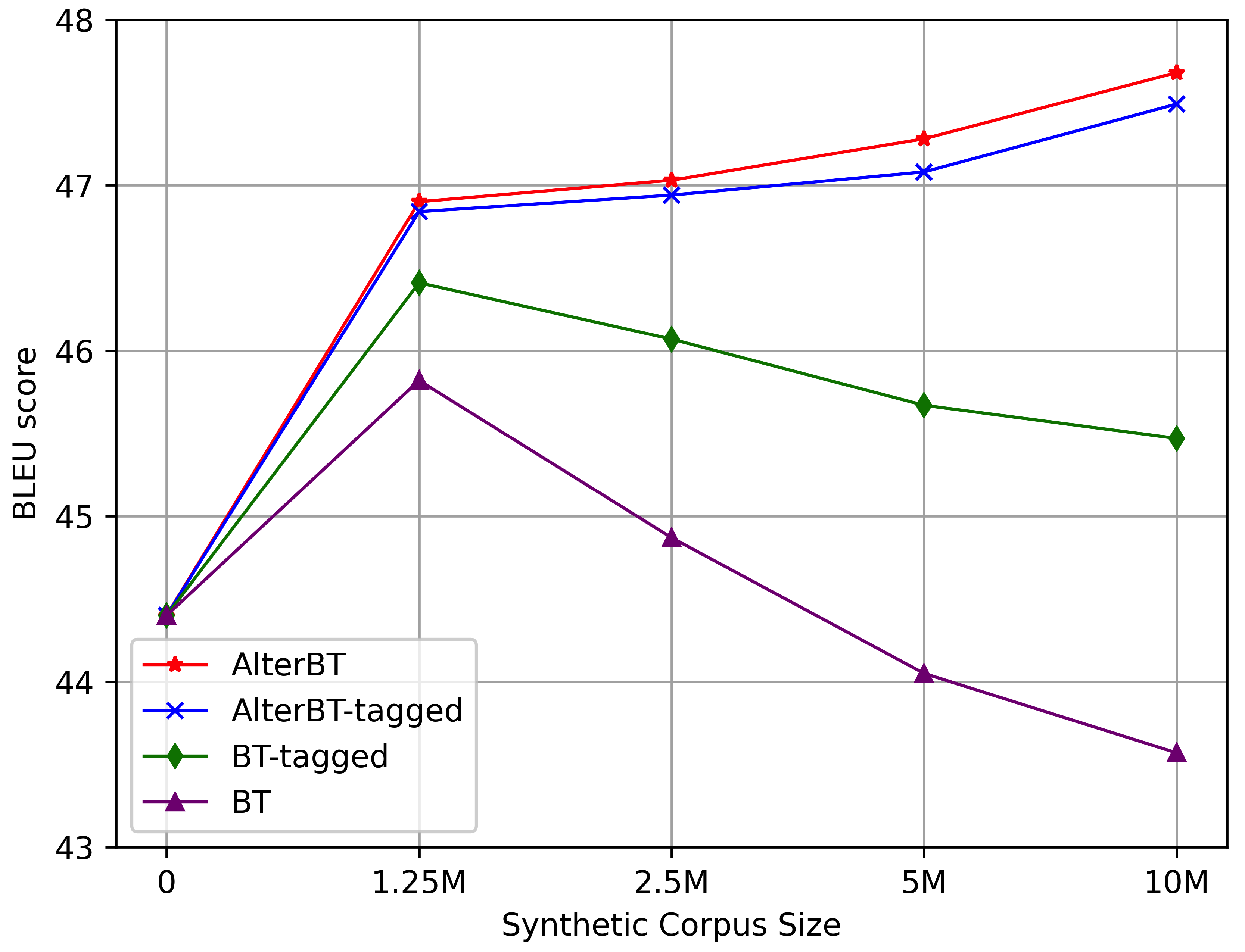}
  \caption{Comparison with several baselines in different data scale. Our alternated approach outperforms the conventional back-translation method and improves the performance of Tagged BT. Moreover, with the enlargement of the synthetic data scale, the BLEU score rises steadily by alternated training.}
\label{fig:zh-en-scale}
\end{minipage}\hfill
\end{figure}

Figure~\ref{fig:zh-en-scale} shows the comparison among several approaches in different scales of training sets on the Chinese-English task. The leftmost point is trained on the authentic data, and other points are trained on the combination of authentic and synthetic corpora. The X-axis shows the synthetic data scale ranging from 1.25M (the size of authentic data) to 10M (the full size of the monolingual corpus). The Y-axis shows the BLEU scores of the combined test set. We find that the performance of BT rises firstly but then decreases as more synthetic data is added, which confirms the findings of \citet{wu2019exploiting}. In contrast, our approach achieves consistent improvement with the enlargement of the synthetic data scale.

Table~\ref{zh-en-nist} shows the detailed translation performance on the Chinese-English task when the synthetic data scale is set to 10M. It can be seen that our alternated training strategy outperforms conventional back-translation and tagged back-translation on all test sets.
We find that during training, the S-Steps account for about 73\% of the total training time, and the A-Steps account for 27\%. This finding suggests that our training procedure composes mainly of S-Steps, and moderate A-Steps are efficient to guide the NMT model towards better points, which lead to the improvement of BLEU performance.


\begin{table}[h!]
\centering
\begin{tabular}{l|ll}
\hline
Scale & 1M & 4.5M\\
\hline
\hline
Base & 34.16 & 34.16 \\
BT & 37.36 & 36.30 \\
BT-tagged & 37.65 & 37.42 \\
\hline
AlterBT & \textbf{38.20}$^{+\dag}$ & 38.53$^{+\dag}$ \\
AlterBT-tagged & 37.98$^{+\dag}$ & \textbf{39.19}$^{+\dag}$  \\
\hline
\end{tabular}
\caption{\label{de-en}
BLEU scores on the IWSLT14 German-English task with 1M and 4.5M additional synthetic corpus. ``$+$" means significantly better than BT (p $<$ 0.01).``\dag" means significantly better than BT-tagged (p $<$ 0.01).}
\end{table}

Table~\ref{de-en} shows the results of the German-English task. Similar to the Chinese-English task, we vary the synthetic data scale from 1M to 4.5M for experiments. We find that the performance degradation also occurs while utilizing large-scale synthetic data, and alternated training approach alleviate the problem and perform better than corresponding baselines.

\subsection{BLEU Landscape Visualization}

To validate the assumption that the authentic data helps to rectify the deviation in synthetic data and redirect the NMT model parameters to a better optimization path, we further investigate the BLEU landscape to compare our method with the BT approach during the same training steps.

\begin{figure}[t]
\centering
\begin{minipage}{1\columnwidth}
 \includegraphics[width=\columnwidth]{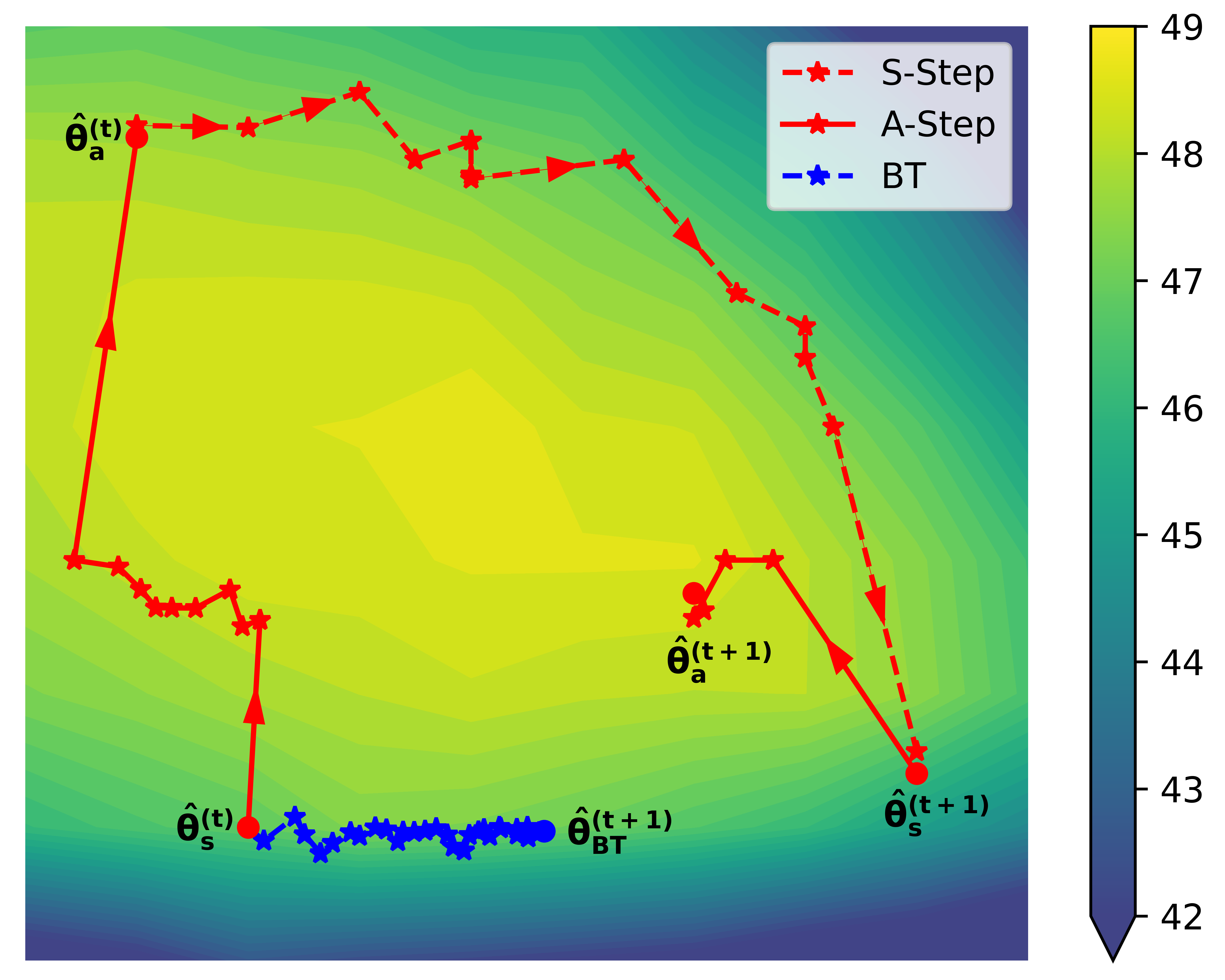}
  \caption{Visualization of BLEU landscape on NIST06 dataset defined by $\hat{\bm{\theta}}_{s}^{(t)}$, $\hat{\bm{\theta}}_{a}^{(t)}$ and $\hat{\bm{\theta}}_{s}^{(t+1)}$. The projected checkpoints are represented as stars. Starting from $\hat{\bm{\theta}}_{s}^{(t)}$, the red dashed and solid segments represent S-Step and A-Step in our method, respectively. The blue dashed segments illustrate the conventional BT method, which shares the same starting point $\hat{\bm{\theta}}_{s}^{(t)}$ with ours. $\hat{\bm{\theta}}_{\mathrm{BT}}^{(t+1)}$ denotes the BT model trained the same steps as $\hat{\bm{\theta}}_{a}^{(t+1)}$. It is shown that alternated training guides the model from $\hat{\bm{\theta}}_{s}^{(t)}$ to $\hat{\bm{\theta}}_{a}^{(t)}$, $\hat{\bm{\theta}}_{s}^{(t+1)}$ and $\hat{\bm{\theta}}_{a}^{(t+1)}$ successively, which finally leads to a better point with a higher BLEU score.} 
\label{fig:bleu-landscape}
\end{minipage}\hfill
\end{figure}

The visualization of the BLEU landscape is shown in Figure~\ref{fig:bleu-landscape}. Checkpoints during alternated training are projected onto the 2D plane 
defined by $\hat{\bm{\theta}}_{s}^{(t)}$, $\hat{\bm{\theta}}_{a}^{(t)}$ and $\hat{\bm{\theta}}_{s}^{(t+1)}$ \footnote{We select $t=2$ for this visualization, and similar performance can be observed for other $t$'s.}.
Our projection method considers both the model parameters and their translation performance (See Appendix~\ref{sec:visualization} for details). For the conventional BT approach, the model parameters are stuck in an inefficient optimization path (highlighted in blue dashed lines). 
In our approach, we find that authentic data effectively guides the model towards a better direction for A-Step (highlighted in red solid lines). For S-Step (highlighted in red dashed lines), although training with synthetic data deteriorates the BLEU performance, it pushes the model away from the original route, and enables authentic data to further redirect the model into a better point with a higher BLEU score.

\section{Related Work}

Our work is based on back-translation (BT), an approach to leverage monolingual data by an additional target-to-source system. 
BT was proved to be effective in neural machine translation (NMT) systems \citep{DBLP:conf/acl/SennrichHB16}.
Despite its effectiveness, BT is limited by the accuracy of synthetic data. Noise and translation errors hinder the boosting of model performance \citep{DBLP:conf/aclnmt/HoangKHC18}. The negative results become more evident when more synthetic data is mixed into training data \citep{DBLP:conf/wmt/CaswellCG19,wu2019exploiting}.

Considerable studies have focused on the accuracy problem in synthetic data and further extended back-translation. \citet{DBLP:conf/aclnmt/ImamuraFS18} demonstrate that generating source sentences via sampling 
increases the diversity of synthetic data and benefits the BT system. \citet{edunov2018understanding} further propose a noisy beam search method to generate more diversified source-side data. \citet{DBLP:conf/wmt/CaswellCG19} add a reserved token to synthetic source-side sentences 
in order to help NMT model distinguish between authentic and synthetic data. Another perspective aims at measuring the translation quality of synthetic data. 
\citet{DBLP:conf/aclnmt/ImamuraFS18} filter sentence pairs with low likelihood or low confidence. \citet{DBLP:conf/emnlp/WangLWLS19} use uncertainty-based confidence to measure words and sentences in synthetic data. Different from the aforementioned works, our approach introduces neither data modification (e.g. noising or tagging) 
nor additional models for evaluation. We alternate training set on the original authentic and synthetic data.

The work relatively close to ours is Iterative Back-Translation \citep{DBLP:conf/aclnmt/HoangKHC18}, which refines forward and backward model via back-translation data, and regenerates more accurate synthetic data from monolingual data. Our work differs from Iterative BT in that we do not require source-side monolingual corpora or repeatedly finetune the backward model.

\section{Conclusion}

In this work, we propose alternated training with synthetic and authentic data for neural machine translation. Experiments have shown the supremacy of our approach. Visualization of the BLEU landscape indicates that alternated training guides the NMT model towards better points.

\section*{Acknowledgments}

This work was supported by the National Key R\&D Program of China (No. 2017YFB0202204), National Natural Science Foundation of China (No.61925601, No.61772302). We thank all anonymous reviewers for their valuable comments and suggestions on this work.

\bibliographystyle{acl_natbib}
\bibliography{anthology,acl2021}

\clearpage

\appendix

\section{Method for Visualization}
\label{sec:visualization}

We first define the projection plane $S$ by parameters $\hat{\bm{\theta}}_{s}^{(t)}$, $\hat{\bm{\theta}}_{a}^{(t)}$ and $\hat{\bm{\theta}}_{s}^{(t+1)}$. Selecting $\hat{\bm{\theta}}^* =\hat{\bm{\theta}}_{s}^{(t)}$ as the basic point and $\boldsymbol{\delta} = \hat{\bm{\theta}}_{a}^{(t)} - \hat{\bm{\theta}}_{s}^{(t)}, \boldsymbol{\eta} = \hat{\bm{\theta}}_{s}^{(t+1)} - \hat{\bm{\theta}}_{s}^{(t)}$ as two basis vectors, we plot the function $f(x,y) = \mathrm{BLEU}(D_{\mathrm{DEV}}; \hat{\bm{\theta}}^* + x\boldsymbol{\delta} + y\boldsymbol{\eta})$ in the 2D surface. We 
calculate the BLEU scores for all NMT models defined by grid points on the projection plane, and construct the BLEU contours via linear interpolation in \textsc{Matplotlib} \citep{matplotlib}.

We project the model checkpoints onto the 2D plane $S$ to represent the parameter geometry and their translation performance. As the 2D contour plane consists of several regions corresponded with different BLEU ranges, we formulate the visualization task into the following problem:
\begin{equation}
    (x_i,y_i) = \argmin_{(x,y) \in S_i}{\left\|\hat{\bm{\theta}}_i - \left (\hat{\bm{\theta}}^* + x\boldsymbol{\delta} + y\boldsymbol{\eta} \right) \right\|_F^2},
\end{equation}
where $S_i$ denotes the BLEU region that the performance of $\hat{\bm{\theta}}_i$ lies in. It is noted that according to the Pythagorean theorem, for $\forall ~(\tilde{x}, \tilde{y}) \in S_i$,
\begin{equation}
    \begin{aligned}
        & \left\|\hat{\bm{\theta}}_i - \left(\hat{\bm{\theta}}^* + \tilde{x}\boldsymbol{\delta} + \tilde{y}\boldsymbol{\eta} \right) \right\|_F^2 \\
        = & \left\|\hat{\bm{\theta}}_i - \left(\hat{\bm{\theta}}^* + \hat{x}_i\boldsymbol{\delta} + \hat{y}_i\boldsymbol{\eta} \right) \right\|_F^2 + \\ 
        & \left\|\left(\hat{\bm{\theta}}^* + \hat{x}_i\boldsymbol{\delta}
        + \hat{y}_i\boldsymbol{\eta} \right) - \left(\hat{\bm{\theta}}^* + \tilde{x}\boldsymbol{\delta} + \tilde{y}\boldsymbol{\eta} \right) \right\|_F^2,    
    \end{aligned}
    \label{pytha}
\end{equation}
where
\begin{equation}
    (\hat{x}_i,\hat{y}_i) = \argmin_{(x,y) \in S}{\left\|\hat{\bm{\theta}}_i - \left (\hat{\bm{\theta}}^* + x\boldsymbol{\delta} + y\boldsymbol{\eta} \right) \right\|_F^2}.
    \label{2d-proj}
\end{equation}

As $(x_i,y_i) \in S_i$, we can substitute $(\tilde{x}, \tilde{y})$ in Eq. (\ref{pytha}) with $(x_i,y_i)$.
Notice that the first term on the right-hand side of Eq. (\ref{pytha}) is independent of $(x_i,y_i)$, the minimizer $(x_i,y_i)$ thus satisfies the following conditions:
\begin{equation}
    \begin{aligned}
        (x_i,y_i) = \argmin_{(\tilde{x},\tilde{y}) \in S_i} \Bigg\| & \left(\hat{\bm{\theta}}^* + \hat{x}_i\boldsymbol{\delta}
        + \hat{y}_i\boldsymbol{\eta} \right) \\
        & - \left(\hat{\bm{\theta}}^* + \tilde{x}\boldsymbol{\delta} + \tilde{y}\boldsymbol{\eta} \right) \Bigg\|_F^2,
    \end{aligned}
    \label{bleu-proj}
\end{equation}
with $(\hat{x}_i,\hat{y}_i)$ satisfying Eq. (\ref{2d-proj}).

According to Eq. (\ref{pytha}), our projection method can be divided into two steps. The first step is to calculate $(\hat{x}_i,\hat{y}_i)$ in Eq. (\ref{2d-proj}), which minimizes the first term of Eq. (\ref{pytha}). By the least square method, we obtain the analytic solution to $(\hat{x}_i,\hat{y}_i)$ as follows:
\begin{equation}
    \begin{cases}
        x_i &= \frac{VB-UC}{B^2-AC}, \\
        y_i &= \frac{UB-VA}{B^2-AC},
    \end{cases}
\end{equation}
where
\begin{equation}
    \begin{aligned}
    A & = \langle \boldsymbol{\delta},\boldsymbol{\delta} \rangle, \\
    B & = \langle\boldsymbol{\delta},\boldsymbol{\eta}\rangle, \\
    C & = \langle\boldsymbol{\eta},\boldsymbol{\eta}\rangle, \\
    U & = \langle\hat{\bm{\theta}}_i - \hat{\bm{\theta}}^*,\boldsymbol{\delta}\rangle, \\
    V & = \langle\hat{\bm{\theta}}_i - \hat{\bm{\theta}}^*,\boldsymbol{\eta}\rangle. \\
    \end{aligned}
\end{equation}
The second step is to solve $(x_i,y_i)$ in Eq. (\ref{bleu-proj}), which minimizes the second term of Eq. (\ref{pytha}). Specially, we have $(x_i,y_i)=(\hat{x}_i,\hat{y}_i)$ if $(\hat{x}_i,\hat{y}_i)\in S_i$. Otherwise, as the BLEU region $S_i$ is enclosed by polygon boundaries with limited edges, we simply calculate the distance between $(\hat{x}_i,\hat{y}_i)$ and each edge and select the minimum one. The boundary point minimizing the distance is then determined as $(x_i,y_i)$. We cast the projection point from $(\hat{x}_i,\hat{y}_i)$ to $(x_i,y_i)$ in order to restore the origin BLEU performance of $\hat{\bm{\theta}}_i$.

\end{document}